\documentclass[lettersize,journal]{IEEEtran}
\usepackage{amsmath,amsfonts}
\usepackage{algorithmic}
\usepackage{algorithm}
\usepackage{array}
\usepackage[caption=false,font=normalsize,labelfont=sf,textfont=sf]{subfig}
\usepackage{textcomp}
\usepackage{stfloats}
\usepackage{url}
\usepackage{verbatim}
\usepackage{graphicx}
\usepackage{cite}
\usepackage{booktabs}
\usepackage[pagebackref,breaklinks,colorlinks,citecolor=cvprblue]{hyperref}
\usepackage[table]{xcolor}
\definecolor{cvprblue}{rgb}{0.21,0.49,0.74}

\begin{document}

\title{Enhancing Perception of Key Changes in Remote Sensing Image Change Captioning}

\author{Cong Yang, Zuchao Li, Hongzan Jiao,~\IEEEmembership{Member}, Zhi Gao,~\IEEEmembership{Member}, Lefei Zhang,~\IEEEmembership{Senior Member}
\thanks{Manuscript received 19 Sep 2024; revised XXX.}
\thanks{The authors are with the National Engineering Research Center for Multimedia Software, School of Computer Science, Wuhan University, Wuhan, 430072, P. R. China, and also with the Hubei Luojia Laboratory, Wuhan, 430079, P. R. China. (e-mail: \{yangcong356, zcli-charlie, zhanglefei\}@whu.edu.cn)}
\thanks{Hongzan Jiao is with the School of Urban Design, Wuhan University, Wuhan 430072, China (e-mail: jiaohongzan@whu.edu.cn).}
\thanks{Zhi Gao is with the School of Remote Sensing and Information Engineering, Wuhan University, Wuhan 430079, China (e-mail: gaozhinus@gmail.com).}}

\markboth{Journal of \LaTeX\ Class Files,~Vol.~14, No.~8, August~2021}%
{Shell \MakeLowercase{\textit{et al.}}: A Sample Article Using IEEEtran.cls for IEEE Journals}


\maketitle

\begin{abstract}
Recently, while significant progress has been made in remote sensing image change captioning, existing methods fail to filter out areas unrelated to actual changes, making models susceptible to irrelevant features. In this article, we propose a novel multimodal framework for remote sensing image change captioning, guided by \textbf{K}ey \textbf{C}hange \textbf{F}eatures and \textbf{I}nstruction-tuned (KCFI). This framework aims to fully leverage the intrinsic knowledge of large language models through visual instructions and enhance the effectiveness and accuracy of change features using pixel-level change detection tasks. Specifically, KCFI includes a ViTs encoder for extracting bi-temporal remote sensing image features, a key feature perceiver for identifying critical change areas, a pixel-level change detection decoder to constrain key change features, and an instruction-tuned decoder based on a large language model. Moreover, to ensure that change description and change detection tasks are jointly optimized, we employ a dynamic weight-averaging strategy to balance the losses between the two tasks. We also explore various feature combinations for visual fine-tuning instructions and demonstrate that using only key change features to guide the large language model is the optimal choice. To validate the effectiveness of our approach, we compare it against several state-of-the-art change captioning methods on the LEVIR-CC dataset, achieving the best performance. Our code will be available at \href{https://github.com/yangcong356/KCFI.git}{https://github.com/yangcong356/KCFI.git}.
\end{abstract}


\section{Introduction}
\IEEEPARstart{I}{mage} change captioning is a new and growing area where natural language processing (NLP) meets computer vision (CV). It focuses on automatically generating detailed descriptions of how a scene has changed between images captured at different times \cite{ZSZ2022}. This process allows machine learning models to better grasp and interpret the evolving world, much like how humans recognize and understand changes in their surroundings. The rapid advancement of remote sensing and Earth observation technologies \cite{ZS2018} has made it easier to obtain remote sensing images of the same area at different times (multi-temporal). As a result, there is growing interest in using multi-temporal remote sensing images for intelligent interpretation. Unlike traditional change detection \cite{LLM2018, TMD2020, SLG2021, ZCZ2024} (CD) methods, which only produce binary images showing changes in geographic areas, remote sensing image change captioning (RSICC) can automatically generate textual descriptions of specific changes in these areas. This capability helps government agencies quickly formulate policies.

\begin{figure}
    \centering
    \includegraphics[width=\linewidth]{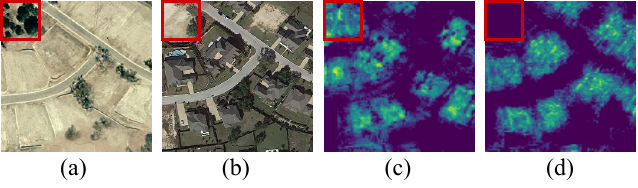}
    \caption{Visualization of irrelevant areas in the change features. (a) Pre-event remote sensing image; (b) Post-event remote sensing image; (c) Visualization of RSICCFormer’s \cite{LZC2022} change features; (d) Visualization of KCFI’s change features. The results in Fig. \ref{ch4-fig3} show that using only change features and textual instructions to guide the large language model in generating change descriptions is the optimal approach. Therefore, filtering out irrelevant change features is crucial for improving the model’s accuracy in change descriptions.}
    \label{ch1-fig1}
\end{figure}

Despite advances in both remote sensing image change detection \cite{DZG2024, HWL2024, CWD2024} and change captioning \cite{CG2023, LZC2023, LYQ2023}, each approach on its own has limitations in providing comprehensive and interactive change explanations. To address this, Liu et al. \cite{LCZ2024} proposed an approach that integrates multi-task learning, combining change detection with change captioning. This approach leverages pixel-level change information, feeding it into a shared image encoder to improve the accuracy of descriptions generated by the change captioning model.

Although incorporating change detection helps the model focus on relevant changes, such as roads and buildings, irrelevant changes (e.g., changes to trees) can distract the model, reducing its overall performance. This challenge is illustrated in Fig. \ref{ch1-fig1}, where bi-temporal remote sensing images show both relevant and irrelevant changes. Moreover, the semantic gap between high-level language descriptions and pixel-level visual details complicates the integration of features from both tasks into a unified model \cite{LZC2023}.

Recent advancements in instruction tuning have shown promise for general-purpose adaptation of large language models (LLMs) \cite{chatgpt, gpt4}. Instruction tuning enables models to use natural language instructions to guide task performance. Drawing inspiration from this, visual instruction tuning has emerged, where language instructions are used to fine-tune large vision models, creating a unified model that handles a wide range of vision-language tasks \cite{ZWJ2024, GGZ2024, CYC2024}. This fine-tuning process aligns visual inputs with language instructions to generate the desired outputs, making it a highly flexible approach for multimodal tasks.

In this paper, we present a novel multimodal framework for RSICC called \textbf{K}ey \textbf{C}hange \textbf{F}eatures and \textbf{I}nstruction-tuned (KCFI). Our approach integrates instruction tuning and leverages the strengths of large language models to generate high-quality textual descriptions of changes in remote sensing images. KCFI focuses on key change features, by using a flow estimation-based module to effectively filter out irrelevant changes and a change detection module to extract meaningful change information. This improves the relevance and accuracy of the descriptions. Specifically, KCFI is composed of four key components: a Vision Transformer (ViT) visual encoder for extracting features from bi-temporal remote sensing images, a key change feature perceiver to focus on key changes, a change detection decoder to improve pixel-level change feature quality, and a large language model fine-tuned with visual instructions for generating detailed change descriptions.

Overall, the main contributions of this paper are as follows:

\begin{enumerate}
    \item This paper proposes an instruction-tuned multimodal framework guided by key change features, leveraging the large language model’s ability to generate high-quality text descriptions. This framework enables the large language model to produce accurate and high-quality change descriptions by utilizing key change features and specific tuning instructions.
    \item We propose a key change feature perception module to identify critical change areas in bi-temporal remote sensing images. Moreover, the key change features extracted by this module are jointly optimized by both semantic-level change descriptions and pixel-level change detection, further enhancing the effectiveness and accuracy of the key change features.
    \item We explore the effect of embedding different combinations of visual features into visual instructions. Through experiments, we demonstrate that when processing visual instructions, inputting only key change features into the large language model is more effective than inputting only bi-temporal remote sensing image features or both bi-temporal image features and key change features.
\end{enumerate}

\section{Related Works}
\subsection{Image Change Detection in Remote Sensing}
Remote sensing image CD is a purely visual perception task, which analyzes and identifies differences in images of the same geographical area captured at different times. Daudt et al. \cite{DSA2018} were the first to introduce an end-to-end trained fully convolutional network (FCN) architecture into change detection. They showed how FCN can boost accuracy and inference speed using an encoder-decoder network and a Siamese architecture with skip connections. This foundational work has paved the way for subsequent advancements in change detection \cite{HWG2023, JWW2023}. Depending on the visual backbone architecture, these methods can be divided into those that rely on convolutional neural networks (CNNs) and those that take advantage of ViTs. To tackle the complex textures and detailed features in high-resolution remote sensing images, Zhang et al. \cite{ZYT2020} proposed a deeply supervised image fusion network (IFN). The IFN utilized a fully convolutional dual-stream architecture and a difference discrimination network (DDN) to extract and fuse multi-level deep features, achieving higher accuracy in change detection and more complete boundary detection. The SNUNet-CD framework \cite{FLS2021} combined a densely connected network structure with an ensemble channel attention module (ECAM) to address the issues of edge pixel uncertainty and small object omission caused by insufficient utilization of shallow positional information in change detection. This SNUNet-CD improved target localization accuracy and overall detection performance. To address the issues of incomplete temporal modeling and spatiotemporal coupling in existing change detection methods, Lin et al. \cite{LYZ2023} proposed the P2V-CD framework, which constructs pseudo-transition videos and employs a decoupled encoder. P2V-CD integrates CNNs and Long Short-Term Memory networks, achieving more accurate detection of temporal and spatial changes.

Due to their superior contextual representation capabilities, the ViTs offer a new approach to change detection by dividing images into patches and using attention mechanisms to capture the global dependencies between these patches. To efficiently model contextual information, Chen et al. \cite{CQS2022} introduced the Bit-Temporal Image Transformer (BIT), which represents bit-temporal images as semantic tokens to capture the high-level concepts of the targeted changes. In cases with limited change sample numbers, SCanFormer \cite{DZG2024} improves change detection accuracy by guiding the learning of semantic features through the joint consideration of spatiotemporal dependencies. TransUNetCD \cite{LZD2022} first uses CNN to model the change features and then uses Transformer to generate a detailed feature map with rich change characteristics, aiming to reduce the redundant information contained in the change features. To ensure the model focuses on both local and global features, Tang et al. \cite{TZM2023} combined CNNs and Transformer, introducing a novel W-shaped dual-branch hierarchical network (WNet) for change detection in remote sensing images. 

\subsection{Image Change Captioning}
Image change captioning is a research area at the intersection of computer vision and natural language processing (NLP) that has gained widespread attention in recent years. This section will briefly review the research progress on generating image change captions in both computer vision and remote sensing fields.

As a pioneering work in image change captioning, Jhamtani et al. \cite{HT2018} were the first to introduce a dataset and corresponding model for this task. The model captured visual salience by introducing a latent variable to align different pixel clusters with the output sentence. Park et al. \cite{PDR2019} emphasized the need to differentiate irrelevant distractions (such as viewpoint changes) from meaningful semantic changes (such as moving objects). To address this, the authors introduced the dual dynamic attention model (DUDA), which uses dual attention to precisely locate changes between 'before' and 'after' images. Kim et al. \cite{KKL2021} proposed the viewpoint-agnostic change captioning network with cycle consistency (VACC), featuring a novel difference encoder to effectively capture true changes and a cycle consistency module to improve model performance. Pointing out that viewpoint changes can obscure semantic differences, Shi et al. \cite{SYG2020} proposed a novel visual encoder designed to distinguish between viewpoint and semantic changes clearly. Additionally, they introduced a reinforcement learning process that fine-tunes the attention mechanism by simulating human attention preferences and using language evaluation rewards to improve caption generation. To generate accurate descriptions, Hosseinzadeh et al. \cite{HW2021} proposed a novel training scheme incorporating composed query image retrieval as an auxiliary task. This approach forced the primary network to produce more detailed captions through additional supervision and introduced a new method for selecting negative candidates to improve performance further. Zheng et al. \cite{ZMS2022} designed a role-playing dialogue system to identify the differences between two similar images. FINER-MLLM \cite{ZWW2024} utilizes LoRA fine-tuning, dual constraints for feature extraction, and retrieval augmentation to detect and describe subtle changes between images effectively.

To describe changes between bi-temporal remote sensing images using human-like sentences, Chouaf et al. \cite{CHS2021} were the first to apply image captioning techniques and proposed a model that combines a CNN with a multimodal RNN to improve the accuracy of change descriptions. Hoxha et al. \cite{HCM2022} and Liu et al. \cite{LZC2022} both focused on change captioning for bi-temporal remote sensing images and contributed by introducing new datasets tailored to this task. While they all choose CNNs as feature extraction, the main distinction was in other architectures: the former employs recurrent neural networks (RNNs) or support vector machines (SVMs) for generating change descriptions, whereas the latter adopts a Transformer-based model featuring a dual-branch Transformer encoder and caption decoder to improve change captioning. Liu et al. \cite{LZC2023} tackled the challenge of confusion in change captioning by separating the task into identifying whether changes occurred and describing those changes. Moreover, they designed a multi-prompt learning strategy that leverages pre-trained LLMs, significantly improving performance. Chang et al. \cite{CG2023} proposed the Chg2Cap network, which integrates a Siamese CNN, an attentive encoder, and a Transformer-based caption generator. Chg2Cap effectively captures significant changes while overcoming illumination and seasonal effects, outperforming existing methods in remote sensing change captioning tasks. To address the inefficiency of visual feature extraction in current RSICC tasks, a single-stream extractor network (SEN) \cite{ZGY2024} pre-trained on bi-temporal remote sensing images was proposed, along with the design of a shallow feature embedding module and a cross-attention guided difference module to improve temporal and difference feature modeling. By introducing a highly efficient state space model, Liu et al. \cite{LCC2024} proposed the RSCaMa model to address the challenge of joint spatial-temporal modeling in remote sensing change captioning. RSCaMa designed a spatial difference-aware state space module for sharp spatial change detection and a temporal-traversing state space module for improved temporal feature interaction, significantly enhancing the efficiency and accuracy of RSICC.

Although the above methods have achieved significant results in remote sensing image change captioning, they fail to suppress the impact of irrelevant areas in the change features. Additionally, existing models do not fully harness the potential of large language models through language instructions. To address this, we propose KCFI, a method that leverages language instructions and key change features to enhance the accuracy of change descriptions.

\subsection{Multi-Modal Large Language Model}
Humans interact through vision and language, each offering unique strengths for understanding the world. This has led to a focus on developing language-augmented vision models that excel in open-world tasks like classification, detection, segmentation, captioning, and visual generation. However, while language bridges visual signals and human communication, its role is often limited to describing image content, restricting the model’s interactivity and adaptability to user instructions.

Recently, instruction tuning has proven highly effective in fine-tuning LLMs for general-purpose use. In instruction tuning, natural language is used to clearly express task instructions and guide end-to-end trainable models to understand and switch between tasks of interest. While models like Alpaca, Vicuna, and GPT-4-LLM have improved LLM alignment by following high-quality natural language instructions, these methods have only been explored with LLMs. Therefore, Liu et al. \cite{LLW2023, LLL2024} proposed visual instruction tuning, which uses language as task instructions to fine-tune large vision models, aiming to build a general-purpose multimodal large language model (MLLM). Since then, visual instruction tuning has been widely explored across various vision tasks. Instruction-ViT \cite{XCY2024} integrated visual instructions for image classification to enhance both performance and adaptability. Lai et al. \cite{LTC2024} proposed a large language-instructed segmentation assistant (LISA), a model that combines the language generation abilities of multimodal LLMs with segmentation capabilities by using an embedding-as-mask approach. Instruct tuning is also utilized in many object detection tasks \cite{WCC2023, WSL2024, DFZ2024} to enhance the perception capability of the model.

Instruction tuning significantly enhances a model’s perception capabilities when tackling more complex image reasoning tasks, allowing it to provide contextual descriptions of images in a human-like manner. GPT4RoI \cite{ZSC2023} introduced spatial instruction tuning for LLMs, enabling more precise interaction with images through both language and bounding boxes on regions of interest (RoI). This innovation allows models to generate fine-grained image captions by aligning language instructions with RoI features. Qwen-VL \cite{BBY2023} leveraged a multi-stage training process and aligned image-caption-box tuples to improve visual and language understanding. These models set new performance records on various visual-language tasks and outperform current vision-language chatbots in real-world dialogue benchmarks. Clever Flamingo \cite{CLD2024} introduced an innovative method for transforming raw vision-language data into visual instruction tuning datasets, effectively reducing the "multimodal alignment tax." Through a U-shaped multi-stage tuning process, the approach focused on improving the model’s instruction-following ability, enhancing its visual comprehension, and refining the politeness of its responses in a structured and efficient way.

\begin{figure*}[!ht]
    \centering
    \includegraphics[width=\linewidth]{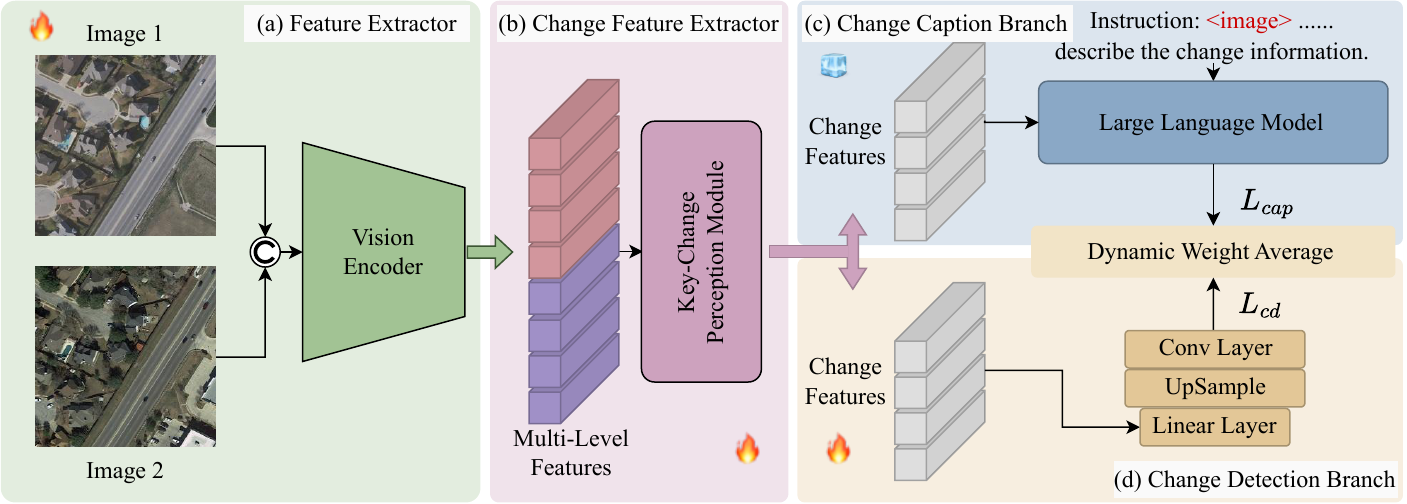}
    \caption{The KCFI framework begins by extracting multi-level features from bi-temporal remote sensing images using a Vision Encoder. By processing these features, the key change perception module can identify key changes. The key change features replace the “\textit{\textless image\textgreater}” token in the Change Caption Branch, where they are used to guide the large language model in generating change descriptions. Simultaneously, the Change Detection Branch processes the change features to generate change masks. A dynamic weight average balances the losses from both the captioning and change detection tasks to ensure optimal performance. The “flames” represent fine-tuning the network parameters, while the “ice cubes” represent freezing the network parameters.}
    \label{ch3-fig1}
\end{figure*}

\section{Methodology}
\subsection{Preliminary}
The proposed KCFI framework consists of a ViT-based feature extractor, a change feature extractor, an LLM decoder, and a change detection decoder, as illustrated in the flowchart in Fig. \ref{ch3-fig1}. During training, the network receives a pair of bi-temporal remote sensing images as input. First, the ViT backbone is responsible for extracting multi-level deep features from the input images. These multi-level features are then passed to the change feature extractor, which includes a specially designed key change feature perception module to precisely locate change information between the two temporal images. Next, The LLM decoder transforms the instructions into word embeddings by substituting the position that corresponds to the \textit{\textless image\textgreater} token in the embedding sequence with the previously extracted key change features, generating natural language descriptions of the image changes. Simultaneously, the change detection branch transforms the change features into change masks to assist in understanding the alterations within the image. Finally, the KCFI framework employs a dynamic weight averaging strategy to balance the caption generation loss and change detection loss, ensuring high accuracy and robustness when performing change captioning tasks.

Formally, we represent a pair of bi-temporal remote sensing images as  $\mathbf{X} = \{ \mathbf{X}_{pre} \in \mathbb{R}^{B \times C \times H \times W}, \mathbf{X}_{post} \in \mathbb{R}^{B \times C \times H \times W} \}$, and the human-annotated text describing the changes between the two images is denoted as  $t_{cap} = \{t_{cap}^i\}_{i=1}^{n}$, where  $n$  is the maximum sentence length. The ground truth annotation of the change mask is denoted as $\mathbf{y} \in \mathbb{R}^{B \times 2 \times H \times W}$. The instruction feed into the large language model is represented as  $t_{ins} = \{\langle \text{image} \rangle, t_{ins}^j\}_{j=1}^{m}$, where “\textit{\textless image\textgreater}” is a special token used to replace the change features, and $m$ is the length of the textual instruction. The KCFI framework takes  $\mathbf{X}$ and $t_{ins}$ as input and outputs the predicted change description and change mask. The entire network is optimized by dynamically adjusting the cross-entropy loss between the predicted change description and the ground-truth text annotation and the binary cross-entropy loss between the predicted change pixels and the actual change pixels.

\subsection{Feature Extractor}
KCFI utilizes a weight-sharing ViT network to extract deep features from the bi-temporal remote sensing images $\mathbf{X}_{pre}$ and $\mathbf{X}_{post}$. The ViT network used in this study is pre-trained on a large-scale dataset and serves as the backbone of our framework. Specifically, the input images are processed by the ViT network, with the layer norm and dropout layers removed, and multi-level features are extracted by capturing features after every four blocks. The resulting deep feature pairs $\{\mathbf{F}_{pre} \in \mathbb{R}^{B \times N \times 4C}, \mathbf{F}_{post} \in \mathbb{R}^{B \times N \times 4C} \}$ represent the multi-level content of the remote sensing images, where $N = \frac{H \times W}{p^2}$ is the sequence length of patches and $p^2$ is the size of each patch. The extraction process is as follows:

\begin{equation}
    \mathbf{F}_{pre}, \mathbf{F}_{post} = f_{\theta}\left( \left[\mathbf{X}_{pre}, \mathbf{X}_{post} \right] \right).
\end{equation}

By utilizing a weight-sharing network, we can efficiently compare the features extracted from the bi-temporal remote sensing images, enabling the effective identification and analysis of differences between the two images in the later stages of the proposed method.

\subsection{Change Feature Extractor}
\begin{figure*}
    \centering
    \includegraphics[width=0.75\linewidth]{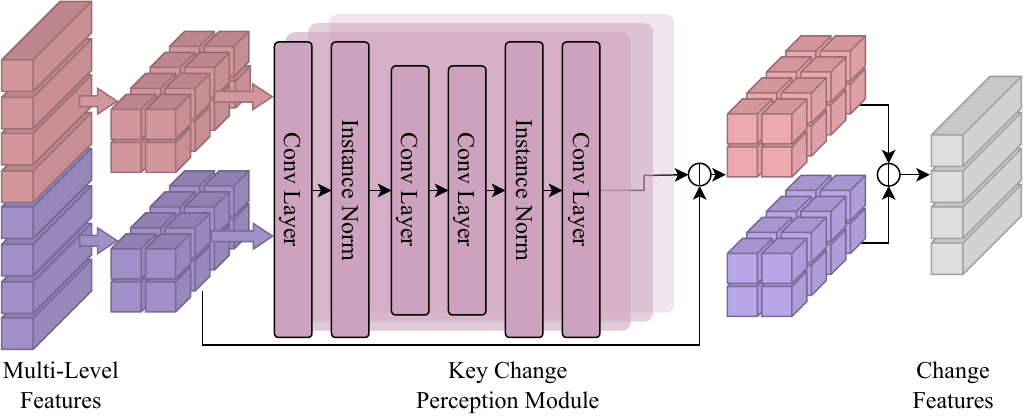}
    \caption{The flowchart of the change feature extractor illustrates the processing flow of multi-level features. First, the features extracted from the four stages are reshaped. Next, these reshaped features are passed in parallel through the key change feature extractor to obtain the final key change features.}
    \label{ch3-fig2}
\end{figure*}

The key-change perception module captures key changes between two input multi-level features by aligning them using estimated flow fields \cite{DFI2015} and fusing the aligned features to emphasize discrepancies. For the bi-temporal multi-level features $\mathbf{F}_{pre}$ and $\mathbf{F}_{post}$ obtained from the feature extractor, we first transform them back into the input format required by the convolutional neural network. Next, we concatenate them along the channel dimension and pass the result through a flow estimation network (as shown in Fig. \ref{ch3-fig2}), resulting in flow fields $\mathbf{F}_1$ and $\mathbf{F}_2$. These flow fields are used to warp the original feature maps via a warping function $\mathcal{W}(\cdot, \cdot)$, producing aligned features:

\begin{equation}
    \begin{aligned}
        \tilde{\mathbf{F}}_{pre} &= \mathcal{W}\left(\mathbf{F}_{pre}, \mathbf{F}_1 \right), \\
        \quad \tilde{\mathbf{F}}_{post} &= \mathcal{W}(\mathbf{F}_{post}, \mathbf{F}_2).
    \end{aligned}
\end{equation}

We then compute residual features by subtracting the original feature maps from the warped ones, highlighting the differences between the aligned features. Finally, these residuals are fused using a specified absolute $\left | \cdot \right |$ difference fusion strategy to produce a feature $\mathbf{F}_{kc}$ that emphasizes the key changes between the inputs. The process of obtaining key change features can be described as follows:

\begin{equation}
    \begin{aligned}
        \mathbf{F}_{pre}^{\prime} &= \tilde{\mathbf{F}}_{pre} - \mathbf{F}_{pre}, \\ \quad \mathbf{x}_{post}^{\prime} &= \tilde{\mathbf{F}}_{post} - \mathbf{F}_{post}, \\
        \mathbf{F}_{kc} &= \left| \mathbf{F}_{pre}^{\prime} - \mathbf{F}_{post}^{\prime} \right|.
    \end{aligned}
\end{equation}

\begin{figure}
    \centering
    \includegraphics[width=0.8\linewidth]{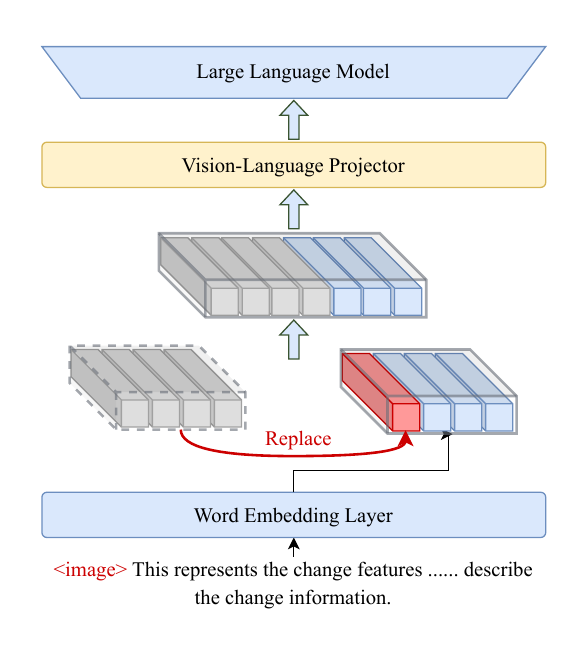}
    \caption{The schematic diagram of the change description branch. First, the instructions are encoded into word vectors, then the “\textit{\textless image\textgreater}” token is replaced with the key change features. The modified features are input into the large language model to generate the change description.}
    \label{ch3-fig3}
\end{figure}

\subsection{Change Captioning Branch}
As shown in Fig. \ref{ch3-fig3}, this study develops a change description decoder based on the Qwen LLM $f_{\phi}$ to generate change descriptions between bi-temporal remote sensing images accurately. Given that the LLM has been pre-trained on a large-scale dataset, we freeze its weights and leverage only key change features and specific instructions to guide it in generating descriptions of changes in bi-temporal remote sensing images. Specifically, before passing the instructions and key change features to the LLM, we first encode these instructions into word vectors $\mathbf{E}_{ins}=\{\langle \text{image} \rangle, e_{ins}^j\}_{j=1}^{m}$ using the model’s embedding layer. Then, we employ a linear projection head $f_{linear}$ to map the key change features into the language space, replacing the “\textit{\textless image\textgreater}” placeholder in the word vectors with the projected feature representations. Finally, these updated word embeddings are fed into the LLM to generate precise textual descriptions of the changes between the bi-temporal remote sensing images. Formally, the calculation for this process is expressed as:

\begin{equation}
    \begin{aligned}
        \mathbf{F}_{v2l} &= f_{linear}\left(\mathbf{F}_{kc}\right), \\
        \mathbf{E}_{new\_ins} &= \{\mathbf{F}_{v2l}, e_{ins}^j\}_{j=1}^{m}, \\
        t_{cap}^{\prime} &= f_{\phi}\left(\mathbf{E}_{new\_ins}\right),
    \end{aligned}
\end{equation}

\noindent where $\mathbf{F}_{v2l}$  represents the key change features projected into the language space, and $t_{cap}^{\prime}$ is the change description output by the large language model.

\subsection{Change Detection Branch}
The change detection branch is designed to generate change detection masks by transforming high-dimensional input features into pixel-wise predictions. It first adjusts the channel dimensions using a linear transformation $f_{linear}^{cd}$ that projects the key change features from high-dimensional to low-dimensional space:

\begin{equation}
    \mathbf{F}^{\prime} = f_{linear}^{cd}\left(\mathbf{F}_{kc}\right),
\end{equation}

\noindent where  $\mathbf{F}_{kc} \in \mathbb{R}^{B \times N \times C_{in}}$  is the input feature tensor,  $B$  is the batch size, and  $N$  is the number of spatial locations (flattened from spatial dimensions). The transformed features  $\mathbf{F}^{\prime}$  are then reshaped and permuted to reconstruct the spatial dimensions  $H^{\prime} \times W^{\prime}$ , where  $H^{\prime} = W^{\prime} = \sqrt{N}$. Next, the feature maps $F^{\prime}$ are upsampled to match the spatial resolution of the original image  $H \times W$  using bilinear interpolation:

\begin{equation}
   \mathbf{F}_{up} = \text{Upsample}(\text{Reshape}\left(\mathbf{F}^{\prime}\right), \text{size}=(H, W)). 
\end{equation}

Finally, a  $1 \times 1$  convolutional layer reduces the channel dimensionality to the number of output classes  $\mathcal{C}$, producing the final segmentation output:

\begin{equation}
    \mathbf{Y} = \text{Conv2D}(\mathbf{F}_{up}),
\end{equation}

\noindent where  $\mathbf{Y} \in \mathbb{R}^{B \times \mathcal{C} \times H \times W}$ . This process effectively transforms the input features into high-resolution, pixel-wise predictions suitable for change detection tasks, allowing for accurate localization of changes at the pixel level.

\subsection{Dynamic Weight Average}
During the training of the KCFI model, we continuously optimize the model parameters through iterative updates within a supervised learning framework to achieve accurate modeling of change captioning and change detection.

A cross-entropy function calculates the difference between the predicted sentence and its corresponding ground truth description in the change description task. The specific description loss function is as follows:

\begin{equation}
    \mathcal{L}_{cap} = -\frac{1}{n} \sum_{i=1}^{n} \sum_{v=1}^{V} t_{cap}^{i,v} \log t_{cap}^{\prime^{i,v}},
\end{equation}

\noindent where $n$ is the total number of word tokens, and $V$ is the size of the vocabulary.  $t_{cap}^{i,v}$  is the one-hot encoding of the ground truth word, where the value is 1 at the index of the correct word and 0 elsewhere. $t_{cap}^{\prime^{i,v}}$ is the model’s predicted probability for the $v$-th word in the vocabulary at time step $i$.

We employ a binary cross-entropy loss function to measure the distinction between the predicted change mask and the ground truth annotations in the change detection task. Specifically, the mathematical definition of this loss function is as follows:

\begin{equation}
    \begin{aligned}
            \mathcal{L}_{det} = -\frac{1}{H \times W} \sum_{i=1}^{H \times W} \left[ \mathbf{y}_i \log(\mathbf{Y}_i)
            + (1 - \mathbf{y}_i) \log(1 - \mathbf{Y}_i) \right],
    \end{aligned}
\end{equation}

\noindent where $\mathbf{y}_i$ is the ground truth label for the $i$-th pixel, and $\mathbf{Y}_i$ is the model’s predicted probability that the $i$-th pixel corresponds to ‘change’.

We adopt a dynamic weight averaging \cite{SEA2019} method to balance the losses between the change captioning and the change detection task. During training, the task weights are dynamically adjusted based on their respective losses, ensuring a balanced contribution from each task:

\begin{equation}
    \begin{aligned}
    &\mathcal{L}_{tot} = \lambda_{1} \mathcal{L}_{cap} + \lambda_{2} \mathcal{L}_{det}, \\
    &\lambda_k(t) = \frac{2 \cdot \exp(w_k(t-1) / T)}{\sum_{i=1}^{2} \exp(w_i(t-1) / T)}, \\
    &w_k(t-1) = \frac{\mathcal{L}_k(t-1)}{\mathcal{L}_k(t-2)}, \\
    &k \in \left \lbrace 1,2\right \rbrace,
    \end{aligned}
\end{equation}

\noindent where $w_k(t-1)$ reflects the improvement speed of task $k$ and $t$ is the training step. We set the temperature parameter $T$ to the commonly used value of 0.2, which is used to control the degree of weight deviation. $\lambda_k(t)$ represents the dynamic weight of task $k$ at time step $t$.

\section{Experiments}
\subsection{Datasets}
This paper explores the joint optimization of change detection and the latent knowledge of LLMs for RSICC. In the experiments, the change captioning dataset LEVIR-CC \cite{LZC2022} and the change detection dataset LEVIR-CD \cite{CS2020} are used for training and evaluation.

The LEVIR-CD and LEVIR-CC dataset offers change data at the pixel level with change detection masks and semantic insights expressed in descriptive sentences. The dataset contains 10,077 pairs of bi-temporal images, which are all 256 × 256 pixels with a resolution of 0.5 meters per pixel. For each image, there is an accompanying annotation mask and five descriptive captions. The maximum word count for every descriptive caption is 39, and the average word count is 7.99. Our training, validation, and test sets contain 6,815, 1,333, and 1,929 pairs of images through the dataset-splitting method detailed in \cite{LZC2022}.

\begin{table}[!ht]
\centering
\caption{Five different instructions designed to guide the LLM}
\label{ch4-tab1}
\begin{tabular}{p{8cm}}
\toprule
\textit{\textless image\textgreater} This represents the change features of geographic targets extracted from remote sensing images. Does this feature contain any information about changes in the geographic targets? If so, please describe the change information. \\ 
\textit{\textless image\textgreater} These are the change characteristics of geographic targets derived from remote sensing images. Do these features reflect any changes in the geographic targets? If yes, please provide details on the changes. \\
\textit{\textless image\textgreater} These are the geographic target change features extracted from remote sensing images. Do these features indicate any changes in the geographic targets? If so, please describe the changes. \\
\textit{\textless image\textgreater} This is the change information of geographic targets extracted from remote sensing images. Does this information reveal any changes in the geographic targets? If so, please describe the changes. \\
\textit{\textless image\textgreater} Here are the change features of geographic targets pulled from remote sensing images. Do these features suggest any changes in the geographic targets? If they do, please describe the nature of those changes. \\
\bottomrule
\end{tabular}
\end{table}

\begin{table*}[!ht]
    \centering
    \caption{Performance comparison with other state-of-the-art methods on the LEVIR-CC Dataset}
    \label{ch4-tab2}
    \begin{tabular}{c c c c c c c c | c}
    \toprule
         Method & BLEU-1 & BLEU-2 & BLEU-3 & BLEU-4 & METEOR & ROUGE-L & CIDEr & $S_{m}^{*}$ \\
        \hline
        RSICCFormer \cite{LZC2022} \scriptsize \textcolor{gray}{[TGRS'22]} &   83.09   &   74.32   &   66.66   &   60.44   &   38.76   &   72.63   &   130.00  &   75.45   \\
        Prompt-CC \cite{LZC2023} \scriptsize \textcolor{gray}{[TGRS'23]} &   83.66   &   75.73   &   69.10   &   63.54   &   38.82   &   73.72  &    136.44  &   78.13   \\
        Sen \cite{ZGY2024} \scriptsize \textcolor{gray}{[TGRS'24]} &   85.10   &   77.05   &   70.01   &   64.09   &   39.59   &   74.57   &   136.02  &   78.57   \\
        SFT \cite{SBC2024} \scriptsize \textcolor{gray}{[Arxiv'24]} & 84.56 & 75.87 &   68.64   &   62.87   &   39.93   &   74.69   &   137.05  &   78.63   \\
        Chg2Cap \cite{CG2023} \scriptsize \textcolor{gray}{[TIP'23]} &   86.14   &   \textbf{78.08}   &   70.66   &   64.39   &   \textbf{40.03}   &   75.12   &   136.61   &   79.03  \\
        RsCaMa \cite{LCC2024} \scriptsize \textcolor{gray}{[LGRS'24]} &   85.79   &   77.99   &   \textbf{71.04}   &   65.24   &   39.91   &   75.24   &   136.56  &   79.24   \\
        \rowcolor{gray!20} \textbf{KCFI}    &   \textbf{86.34}   &   77.31   &   70.89   &   \textbf{65.30}   &   39.42   &   \textbf{75.47}   &   \textbf{138.25}  &   \textbf{79.61}   \\
    \bottomrule
    \end{tabular}
\end{table*}

\subsection{Implementation Details}
The method proposed in this study was implemented using Huggingface's transformers framework and trained on four NVIDIA A100 GPUs. During training, the AdamW \cite{LH2019} optimizer is used with a global initial learning rate of 0.0001. The learning rates for the visual encoder and projection layer are both set to 0.00001, and the weight decay was set to 0.0005. The training process concludes after 50 epochs, with a batch size of 32 per GPU. Moreover, we employ a cosine annealing strategy with a warm-up learning rate, using an epoch ratio of 0.03 for the warm-up phase. As shown in Table \ref{ch4-tab1}, we have designed five different instructions to guide the large language model.

\subsection{Evaluation Metrics}
To evaluate the performance of the proposed model in describing changes between bi-temporal images, we use several evaluation metrics commonly employed in previous change captioning studies, including BLEU \cite{PRW2002}, METEOR \cite{BL2005}, ROUGE-L \cite{L2004}, and CIDEr-D \cite{VZP2015}:

\begin{enumerate}
    \item BLEU evaluates how closely the generated text matches the reference text by measuring the overlap of n-grams, from individual words to longer sequences.
    \item METEOR utilizes an external dictionary for synonym recognition and stemming, providing a more thorough evaluation of the quality of the generated text.
    \item ROUGE-L measures the similarity between the generated and reference texts by analyzing the length of their longest common subsequence
    \item CIDEr-D evaluates the diversity and accuracy of generated descriptions by comparing word frequency patterns between the generated text and a set of reference texts. It places particular emphasis on ensuring consistency with multiple reference texts. 
\end{enumerate}

Furthermore, following prior research \cite{ZZY2022}, we use the composite metric $S^{*}_{m}$ to integrate these individual metrics:

\begin{equation}\label{eq4-1}
    S_m^*=\frac{1}{4} *\left(\text {BLEU-4}+\text {METEOR}+\text{ROUGE-L}+\text{CIDEr-D}\right)
\end{equation}




\subsection{Comparison Methods}
We benchmark six state-of-the-art RSICC methods on the LEVIR-CC dataset, including RSICCFormer, Prompt-CC, Sen, SFT, Chg2Cap, and RsCaMa. A more detailed introduction to these methods is provided below.

\begin{enumerate}
    \item RSICCFormer \cite{LZC2022} is composed of three main components: 1) a CNN-based feature extractor that generates high-level features of remote sensing image pairs; 2) a dual-branch Transformer encoder designed to enhance feature discrimination; and 3) a caption decoder that generates sentences describing the changes.
    \item Prompt-CC \cite{LZC2023} designs an image-level classifier to identify whether a change has occurred in the image and uses a feature-level decoder to extract discriminative features to determine the specific change. Moreover, Prompt-CC leverages prompt learning, utilizing LLMs to generate change captions for remote sensing images.
    \item Sen \cite{ZGY2024} pre-trains a single-stream feature extractor on bi-temporal remote sensing images using contrastive learning to reduce the domain gap. Furthermore, Sen introduces a shallow feature embedding module and a cross-attention guided difference module to improve the feature modeling of difference information.
    \item SFT \cite{SBC2024} consists of three components: a CNN-based high-level feature extractor, a sparse focus attention mechanism-based transformer encoder to locate and capture change regions in bi-temporal images, and a description decoder that embeds image and text data to generate sentences describing the changes.
    \item Chg2Cap \cite{CG2023} utilizes an attentive encoder and a Transformer-based decoder for change captioning in remote sensing images. The encoder captures internal and external information through hierarchical self-attention blocks and enhances consistent and inconsistent features using residual blocks.
    \item RsCaMa \cite{LCC2024} leverages Mamba, a state space model with a global receptive field and linear complexity, to construct the spatial difference-aware state space module and temporal-traversing state space module, enabling efficient spatial-temporal joint modeling and information interaction.
\end{enumerate}

\begin{figure*}[!ht]
    \centering
    \footnotesize
    \begin{tabular}{p{5.6cm} p{5.6cm} p{5.6cm}}
    \includegraphics[width=2.75cm]{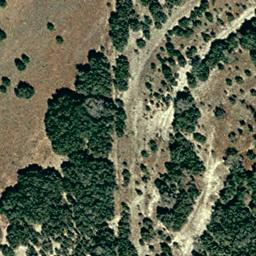} \includegraphics[width=2.75cm]{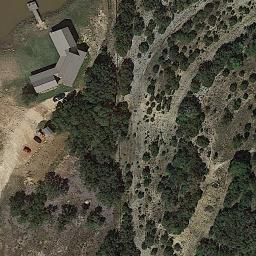} &
    \includegraphics[width=2.75cm]{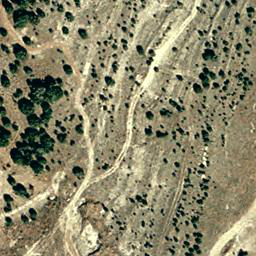} \includegraphics[width=2.75cm]{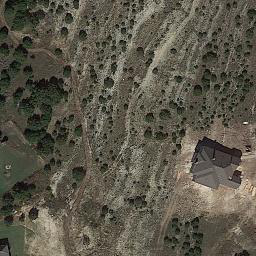} &
    \includegraphics[width=2.75cm]{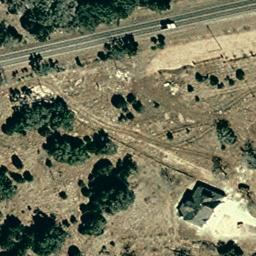} \includegraphics[width=2.75cm]{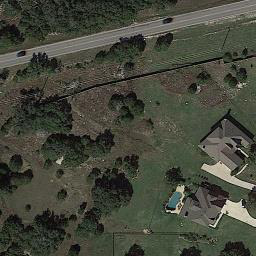} \\
    \textbf{GT:} a building appears at the top and some trees disappear. & \textbf{GT:} a house shows up in the desert. & \textbf{GT:} a new villa is built next to the other one.  \\
    \textbf{RSICCFormer:} some houses are built in the woods. & \textbf{RSICCFormer:} \textcolor{blue}{two} houses are built in the \textcolor{blue}{woods}. & \textbf{RSICCFormer:} some houses are built on the bareland. \\
    \textbf{PromptCC:} a \textcolor{blue}{villa} appears in the upper left corner of the scene. & \textbf{PromptCC:} a house appears in the desert. & \textbf{PromptCC:} a villa is built in the bottom right. \\
    \textbf{Chg2Cap:} some trees are removed and a house is built. & \textbf{Chg2Cap:} a house with a \textcolor{blue}{path} appears in the desert. & \textbf{Chg2Cap:} two buildings appear on the bareland. \\
    \textbf{KCFI:} \textcolor{red}{a building appears in the top left corner} of the scene. & \textbf{KCFI:} \textcolor{red}{a house is built in the desert}. & \textbf{KCFI:} \textcolor{red}{two villas} are built in the \textcolor{blue}{woods}. \\
    \multicolumn{1}{c}{(a)} & \multicolumn{1}{c}{(b)} & \multicolumn{1}{c}{(c)} \\
    \includegraphics[width=2.75cm]{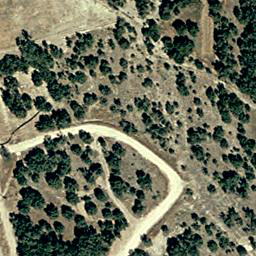} \includegraphics[width=2.75cm]{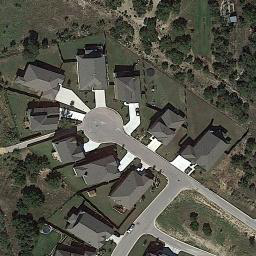} &
    \includegraphics[width=2.75cm]{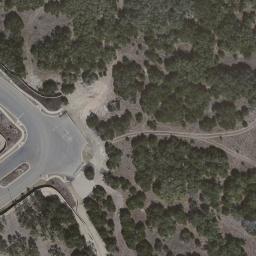} \includegraphics[width=2.75cm]{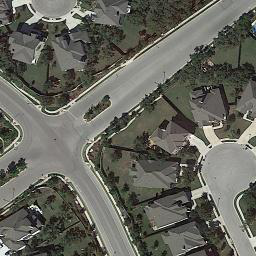} &
    \includegraphics[width=2.75cm]{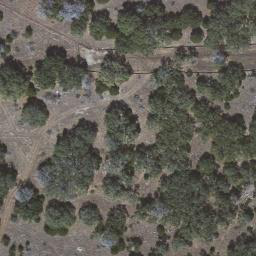} \includegraphics[width=2.75cm]{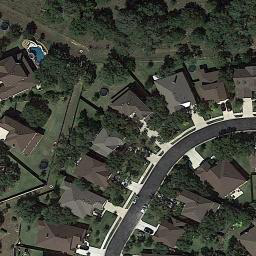} \\
    \textbf{GT:} trees and a road are removed and many villas with a new road are built. & \textbf{GT:} the main road is completed and several buildings are constructed on both sides of the roads. & \textbf{GT:} a winding road has been built across the forest and many villas have been constructed neatly on both sides of it.  \\
    \textbf{RSICCFormer:} many houses are built along the road. & \textbf{RSICCFormer:} many houses are built along the road. & \textbf{RSICCFormer:} a road with many houses built along appears in the \textcolor{blue}{desert}. \\
    \textbf{PromptCC:} the woods are removed and a road with villas built along on both sides of it. & \textbf{PromptCC:} many houses are built on both sides of the road. & \textbf{PromptCC:} a road is built across the forest and many villas are constructed on both sides of it. \\
    \textbf{Chg2Cap:} trees are removed and a road with villas built along and many villas built along the road. & \textbf{Chg2Cap:} a road is built to replace the woods. & \textbf{Chg2Cap:} a road with houses built on both sides appears in the \textcolor{blue}{desert}. \\
    \textbf{KCFI:} trees are removed and a road with villas built along appears. & \textbf{KCFI:} a road is built and many villas are constructed along the road. & \textbf{KCFI:} \textcolor{red}{a winding road} with many villas built along replaces the forest. \\
    \multicolumn{1}{c}{(d)} & \multicolumn{1}{c}{(e)} & \multicolumn{1}{c}{(f)} \\
    \includegraphics[width=2.75cm]{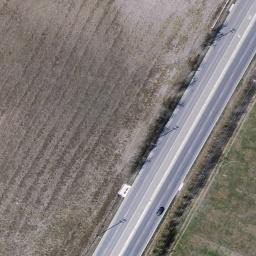} \includegraphics[width=2.75cm]{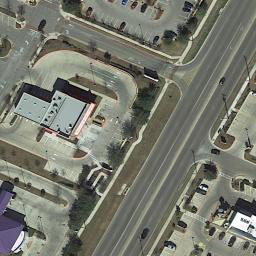} &
    \includegraphics[width=2.75cm]{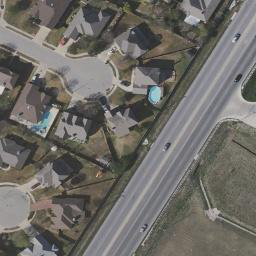} \includegraphics[width=2.75cm]{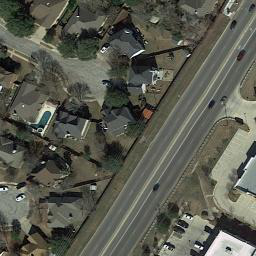} &
    \includegraphics[width=2.75cm]{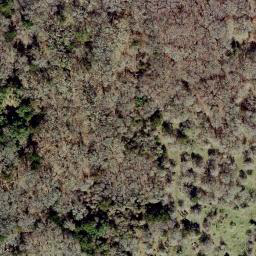} \includegraphics[width=2.75cm]{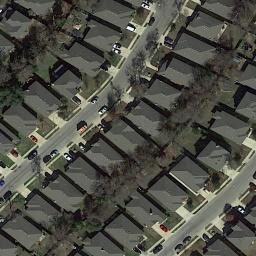} \\
    \textbf{GT:} some buildings and parking lots are constructed beside the road on the bareland. & \textbf{GT:} two buildings and a parking lot are built on one side of the original road. & \textbf{GT:} the woods are developed into neat rows of buildings and two roads in between. \\
    \textbf{RSICCFormer:} some buildings and a parking lot appear on the \textcolor{blue}{bareland}. & \textbf{RSICCFormer:} a road is built on the bareland. & \textbf{RSICCFormer:} a road with rows of houses and roads appears. \\
    \textbf{PromptCC:} a building with a parking lot is built on the bareland. & \textbf{PromptCC:} some houses are built at the \textcolor{red}{bottom right corner} of the scene. & \textbf{PromptCC:} a residential area with many houses and roads appears. \\
    \textbf{Chg2Cap:} some buildings are built near the road. & \textbf{Chg2Cap:} some houses are built at the bottom corner. & \textbf{Chg2Cap:} the forest is replaced by a residential area with rows of houses on both sides  \\
    \textbf{KCFI:} a building and a parking lot appear on the bareland. & \textbf{KCFI:}  \textcolor{red}{two buildings with a parking lot} and some roads appear on the bareland. & \textbf{KCFI:} the forest disappears and two rows of houses are built on both sides of the road. \\
    \multicolumn{1}{c}{(g)} & \multicolumn{1}{c}{(h)} & \multicolumn{1}{c}{(i)} \\
    \end{tabular}
    \caption{Change captioning results generated by RSICCFormer, PromptCC, Chg2Cap, and the KCFI on the LEVIR-CC dataset. \textbf{GT} represents one of the five ground truth annotations from the original dataset. \textcolor{red}{Red} ones indicate accurate descriptions, while those in \textcolor{blue}{blue} represent incorrect descriptions.}
    \label{ch4-fig1}
\end{figure*}

\subsection{Quantitative Comparison}
A shown in Table \ref{ch4-tab2} The KCFI method demonstrates superior performance across multiple key metrics, standing out, particularly in BLEU-1 (86.34), BLEU-4 (65.30), ROUGE-L (75.47), CIDEr (138.25), and $S_{m}^{}$ (79.61). It excels in capturing both word-level and sentence-level accuracy, as reflected in its top BLEU-1 and BLEU-4 scores, highlighting its ability to generate detailed and coherent descriptions that closely match the reference texts. The highest ROUGE-L score further underscores its capability to maintain structural consistency with reference descriptions. In terms of semantic accuracy, the method achieves the best CIDEr score, indicating its strong ability to generate relevant and content-rich descriptions. Its leading $S_{m}^{}$ score demonstrates a clear advantage in identifying significant change regions, which is crucial for remote sensing change description tasks.

While the method slightly underperforms in BLEU-2 and BLEU-3 compared to some competing approaches, and the METEOR score (39.42) suggests room for improvement in sentence diversity and synonymy, these limitations are relatively minor. Overall, the KCFI method generates accurate, content-rich, and structurally consistent descriptions, making it highly effective for RSICC.

\begin{figure*}
    \centering
    \footnotesize
    \begin{tabular}{p{8.5cm} p{8.5cm}}
         \includegraphics[width=2.05cm]{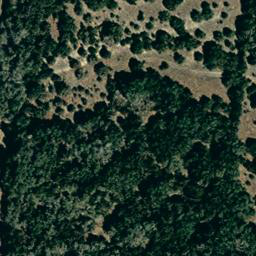} \includegraphics[width=2.05cm]{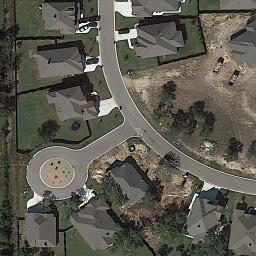} \includegraphics[width=2.05cm]{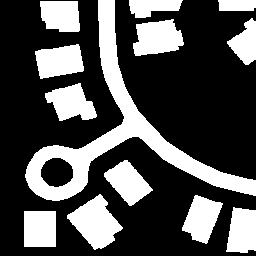} \includegraphics[width=2.05cm]{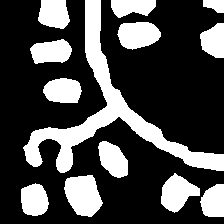} & \includegraphics[width=2.05cm]{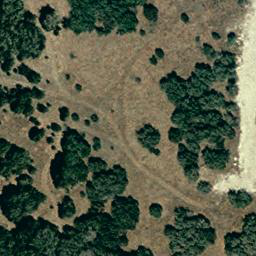} \includegraphics[width=2.05cm]{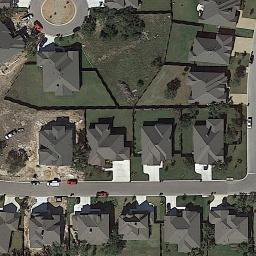} \includegraphics[width=2.05cm]{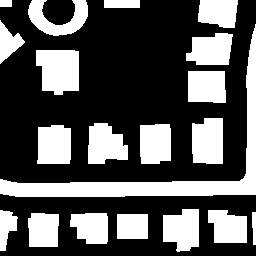} \includegraphics[width=2.05cm]{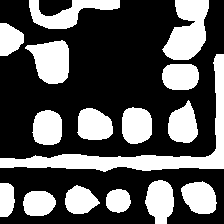}\\ 
         \textbf{GT}: the vegetation has been replaced by a road and many villas around. & \textbf{GT}: a road appears at the bottom and many houses are scattered replacing the trees.\\
         \textbf{KCFI}: a road is built on the bareland and some houses are built along it. & \textbf{KCFI}: many buildings have been built and some roads are constructed in the bareland. \\
         \includegraphics[width=2.05cm]{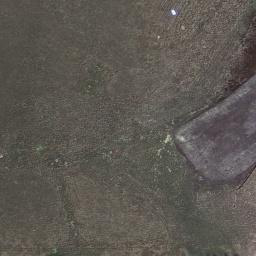} \includegraphics[width=2.05cm]{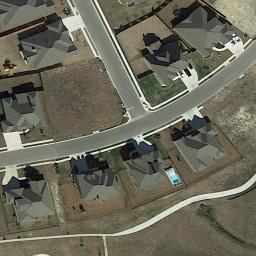} \includegraphics[width=2.05cm]{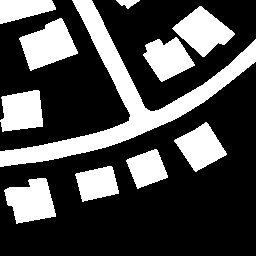} \includegraphics[width=2.05cm]{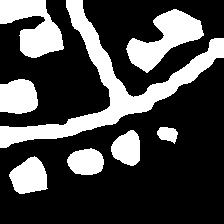} & \includegraphics[width=2.05cm]{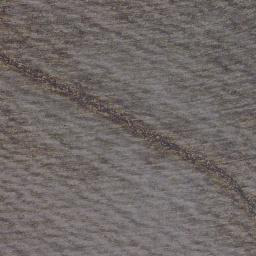} \includegraphics[width=2.05cm]{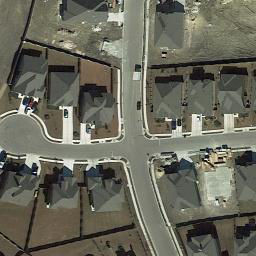} \includegraphics[width=2.05cm]{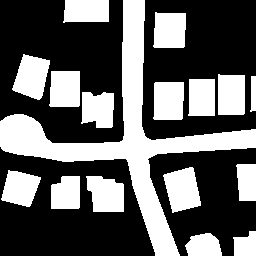} \includegraphics[width=2.05cm]{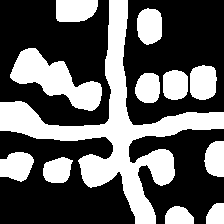}\\ 
         \textbf{GT}: some roads and houses are built on bareland. & \textbf{GT}: some roads and houses are built on bareland. \\
         \textbf{KCFI}: a road with villas is built on the bareland. & \textbf{KCFI}: a road is built on the bareland and many houses are built around it. \\
    \end{tabular}
    \caption{The change detection results of the KCFI model and their corresponding change descriptions.}
    \label{ch4-fig2}
\end{figure*}

\subsection{Qualitative Comparison}
The visualization in Fig. \ref{ch4-fig1} showcases the change captioning results generated by four models—RSICCFormer, PromptCC, Chg2Cap, and KCFI—on the LEVIR-CC dataset, with ground truth annotations (GT) provided for comparison. Red text indicates accurate descriptions, while blue text highlights incorrect ones. Across various scenarios, KCFI consistently demonstrates superior performance by accurately identifying key changes in the scenes, such as specific building and villa placements, while other models either miss important details or provide less relevant descriptions. For example, in (a), KCFI correctly captures "a building appears in the top left corner," closely aligning with the GT, whereas other models fail to mention the exact location or provide imprecise information. In more complex scenes, such as in (e), KCFI accurately identifies both the road construction and the villas built along it, outperforming RSICCFormer and Chg2Cap, which either omit crucial details or misinterpret the changes. Throughout the examples, KCFI's ability to focus on critical change areas and filter out irrelevant details results in more precise and meaningful captions. In contrast, models like PromptCC and Chg2Cap occasionally emphasize less relevant features or misinterpret the context, leading to less accurate descriptions. Overall, KCFI consistently generates captions that are closely aligned with GT annotations, demonstrating the effectiveness of its instruction-tuned, key change feature-guided framework in generating high-quality, accurate change descriptions.

\subsection{Ablation Studies}

\begin{figure}[!ht]
    \centering
    \includegraphics[width=\linewidth]{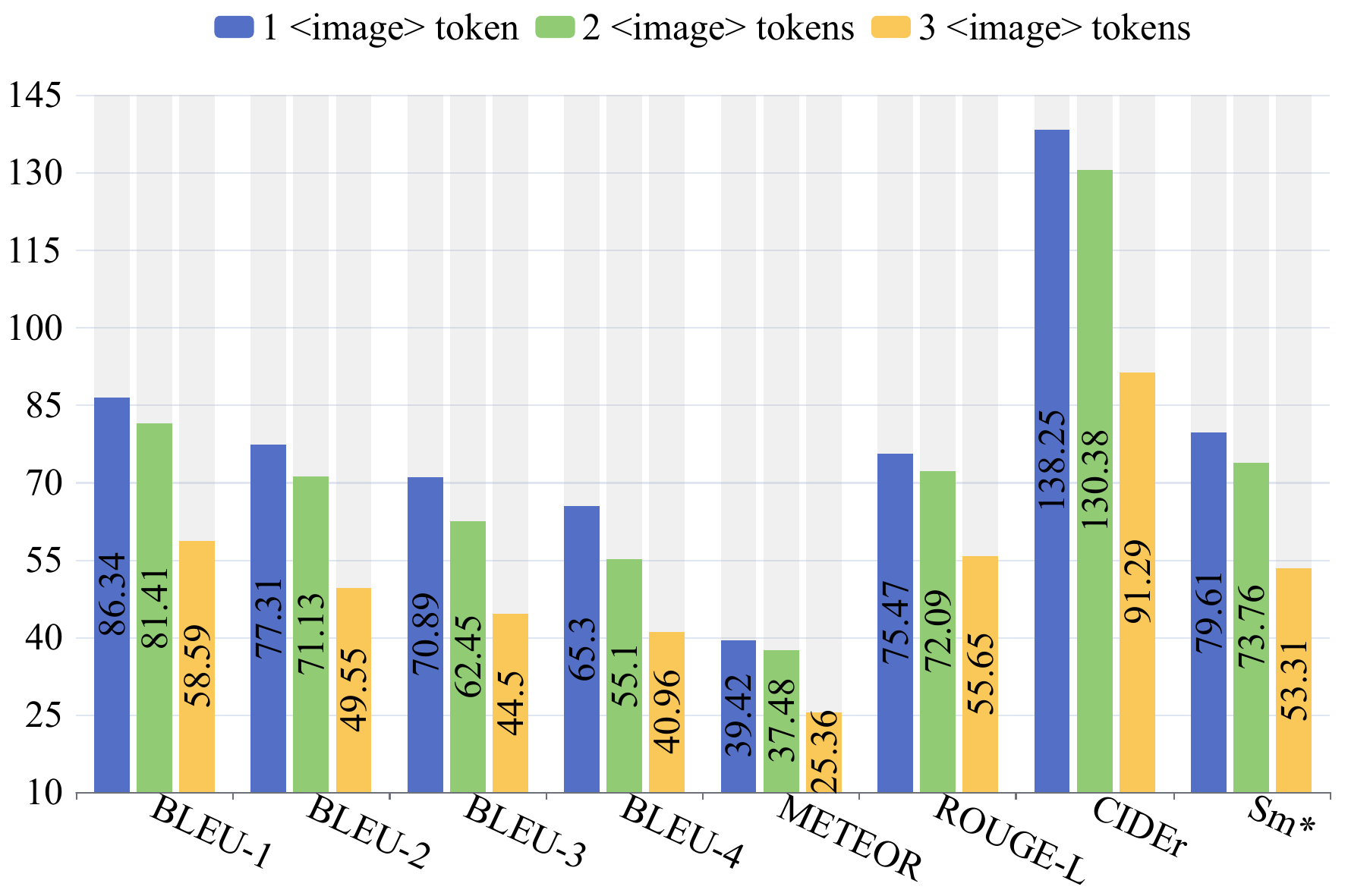}
    \caption{The impact of the number of \textit{\textless image\textgreater} tokens on KCFI performance.}
    \label{ch4-fig3}
\end{figure}

\subsubsection{Use One or Multiple \textit{\textless image\textgreater} Tokens}
We investigate the impact of using different numbers of \textit{\textless image\textgreater} tokens on the task of RSICC. The experiment is configured in three scenarios: “1 \textit{\textless image\textgreater} token” represents using only change features, “2 \textit{\textless image\textgreater} tokens” represents using bi-temporal image features, and “3 \textit{\textless image\textgreater} tokens” represents using both bi-temporal image features and change features. These features are input into a large language model by replacing the \textit{\textless image\textgreater} token in the instruction to generate change captions.

The experimental results in Fig. \ref{ch4-fig3} show that using 1 \textit{\textless image\textgreater} token, i.e., only change features, achieves the best performance across all metrics. Specifically, BLEU-1 to BLEU-4 scores reach 86.34, 77.31, 70.89, and 65.30, respectively, indicating that this configuration effectively captures the key change information between bi-temporal images, generating descriptions that are closer to the reference captions. The CIDEr score of 138.25 further confirms that the model strikes a good balance between diversity and consistency in the descriptions. In contrast, using 2 \textit{\textless image\textgreater} tokens, which corresponds to bi-temporal image features alone, results in weaker performance, particularly with BLEU-4 and CIDEr scores of 55.10 and 130.38, respectively. This suggests that relying solely on bi-temporal image features makes it difficult to distinguish subtle changes between images.

When using 3 \textit{\textless image\textgreater} tokens, which combine both bi-temporal image features and change features, the overall performance declines significantly. In particular, the BLEU scores drop sharply, which could be due to the introduction of too many details, making it harder for the large language model to focus on the primary changes. The METEOR and ROUGE-L metrics also reflect this trend, with the 1 \textit{\textless image\textgreater} token model achieving the highest scores of 39.42 and 75.47, respectively, while the models using 2 or 3 \textit{\textless image\textgreater} tokens perform worse. This suggests that as the number of \textit{\textless image\textgreater} tokens increases, the generated descriptions may include more irrelevant details, negatively affecting semantic consistency with the reference descriptions.

The experimental results indicate that using 1 \textit{\textless image\textgreater} token, i.e., only change features, significantly enhances model performance in the task of RSICC. In contrast, using 2 or 3 \textit{\textless image\textgreater} tokens, while introducing more image features, also introduces excessive irrelevant information, thereby weakening the model’s ability to effectively capture key changes.

\subsubsection{Impact of Fine-Tuning Visual Encoder}
As shown in Table \ref{ch4-tab3}, we compare the impact of fine-tuning versus not fine-tuning the visual encoder on the performance of the RSICC task. The results show that the model with a fine-tuned visual encoder outperforms the non-fine-tuned model across all evaluation metrics, demonstrating that fine-tuning helps the model better understand the changes between bi-temporal images and generate more accurate descriptions.

Specifically, the fine-tuned model shows significant improvements in BLEU-1 to BLEU-4 scores, particularly with BLEU-4 increasing from 53.15 to 65.30, indicating enhanced consistency in longer n-gram sequences. The fine-tuned model also achieves higher scores in the METEOR and ROUGE-L metrics, highlighting its superior ability to capture morphological variations and semantic similarity. Additionally, the increase in CIDEr scores suggests that the fine-tuned model generates more diverse and richer descriptions. The improvement in the $S_{m}^{*}$ composite metric further validates the effectiveness of the fine-tuning strategy.

\begin{table*}[!ht]
    \centering
    \caption{Exploring the impact of fine-tuning the visual encoder on performance using the LEVIR-CC dataset}
    \begin{tabular}{c c c c c c c c | c}
    \toprule
          & BLEU-1 & BLEU-2 & BLEU-3 & BLEU-4 & METEOR & ROUGE-L & CIDEr & $S_{m}^{*}$\\
        \hline
        \textbf{KCFI} \scriptsize{w/o ft}    &   80.67   &   69.87   &  60.86   &   53.15   &   37.07   &   72.17   &   128.64  &   72.75 \\
        \rowcolor{gray!20} \textbf{KCFI} \scriptsize{w/ ft}    &   \textbf{86.34}   &   \textbf{77.31}   &  \textbf{70.89}   &   \textbf{65.30}   &   \textbf{39.42}   &   \textbf{75.47}   &   \textbf{138.25}  &   \textbf{79.61}  \\
    \bottomrule
    \multicolumn{9}{l}{"w/o ft" and "w/ ft" represent without fine-tuning and with fine-tuning, respectively.}
    \end{tabular}
    \label{ch4-tab3}
\end{table*}

\begin{table*}[!ht]
    \centering
    \caption{Experiments validating the effectiveness of different modules and strategies on the LEVIR-CC dataset}
    \begin{tabular}{c c c c | c c c c c c c | c }
    \toprule
        Baseline & CFE & CDH & DWA  & BLEU-1 & BLEU-2 & BLEU-3 & BLEU-4 & METEOR & ROUGE-L & CIDEr & $S_{m}^{*}$\\
        \hline
        \checkmark & - & - & - &   81.41   &  71.13  &  62.45  &  55.10  &  37.48  &  72.09  &  130.38  &  73.76  \\
        \checkmark & \checkmark & - & - & 85.07 & 75.18 & 66.29 & 58.84 & 38.71 & 74.20 & 136.17 & 76.89 \\
        \checkmark & \checkmark & \checkmark & - & 86.27 & 77.25 & 68.85 & 61.37 & 39.36 & 75.36 & 137.08 & 78.29 \\
        \rowcolor{gray!20} \checkmark & \checkmark & \checkmark & \checkmark  &  \textbf{86.34}   &   \textbf{77.31}   &  \textbf{70.89}   &   \textbf{65.30}   &   \textbf{39.42}   &   \textbf{75.47}   &   \textbf{138.25}  &   \textbf{79.61} \\
    \bottomrule
    \end{tabular}
    \label{ch4-tab4}
\end{table*}

\subsubsection{The Role of Each Module and Strategy}
Based on the ablation study results in Table \ref{ch4-tab4}, we systematically evaluate the impact of each module and strategy on model performance. In the experiment, the ‘baseline’ model is used as a control, which only integrated bi-temporal remote sensing image features and applied textual instructions. In other words, the ‘baseline’ model corresponds to the case in Fig. \ref{ch4-fig3} where only one \textit{\textless image\textgreater} token is used. We then gradually introduce the change feature extractor (CFE), change detection branch (CDH), and dynamic weight averaging (DWA) strategy to validate their effectiveness in the RSICC task.

The Baseline model demonstrates basic performance with BLEU-1 (81.41), METEOR (37.48), and CIDEr (130.38) scores. However, since it only relies on bi-temporal image information, there is room for improvement in generating accurate change descriptions. With the addition of the CFE module, the model shows a significant improvement in BLEU-4 (from 55.10 to 58.84) and CIDEr (from 130.38 to 136.17), indicating that CFE effectively enhances the model's ability to extract change features. Further introducing the CDH module leads to additional gains, raising BLEU-4 to 61.37 and CIDEr to 137.08, proving that a dedicated change detection branch improves the precision of change region localization.

Finally, including the DWA strategy brings the model to its best performance. By dynamically balancing the caption loss and change detection loss, the model achieves a better balance between description generation and change detection. All evaluation metrics see significant improvements, with BLEU-4 reaching 65.30, CIDEr increasing to 138.25, and $S_{m}^{*}$ rising from 78.29 to 79.61. This further validates DWA's role in enhancing the model's overall performance.

The gradual introduction of the CFE, CDH, and DWA modules significantly boosts the model's performance in the RSICC task, particularly in generating more accurate and semantically rich descriptions. The experimental results demonstrate that the combined use of these modules enables the model to achieve notable progress in capturing image changes and generating high-quality descriptions.

\section{Conclusion}
We propose a key change features-guided instruction-tuned multimodal framework for remote sensing image change captioning. This approach uses a key change perception module to filter out unimportant change areas, enabling the extraction of critical change features from bi-temporal remote sensing images. Additionally, we design various visual fine-tuning instructions to guide the large language model in generating descriptions of key change areas. We also leverage a pixel-level change detection task to enhance the effectiveness and accuracy of the key change features, further improving the framework’s precision in describing change areas in bi-temporal images. Compared to existing state-of-the-art methods, KCFI demonstrates clear advantages. Furthermore, we explore the performance of different features as visual guidance instructions and confirm the superiority of using key change features combined with textual instructions to generate change descriptions.

\bibliographystyle{IEEEtran}
\bibliography{Bibtex/ref.bib}

\begin{thebibliography}{10}
\providecommand{\url}[1]{#1}
\csname url@samestyle\endcsname
\providecommand{\newblock}{\relax}
\providecommand{\bibinfo}[2]{#2}
\providecommand{\BIBentrySTDinterwordspacing}{\spaceskip=0pt\relax}
\providecommand{\BIBentryALTinterwordstretchfactor}{4}
\providecommand{\BIBentryALTinterwordspacing}{\spaceskip=\fontdimen2\font plus
\BIBentryALTinterwordstretchfactor\fontdimen3\font minus \fontdimen4\font\relax}
\providecommand{\BIBforeignlanguage}[2]{{%
\expandafter\ifx\csname l@#1\endcsname\relax
\typeout{** WARNING: IEEEtran.bst: No hyphenation pattern has been}%
\typeout{** loaded for the language `#1'. Using the pattern for}%
\typeout{** the default language instead.}%
\else
\language=\csname l@#1\endcsname
\fi
#2}}
\providecommand{\BIBdecl}{\relax}
\BIBdecl

\bibitem{CWD2024}
H.~Chang, P.~Wang, W.~Diao, G.~Xu, and X.~Sun, ``Remote sensing change detection with bitemporal and differential feature interactive perception,'' \emph{{IEEE} Trans. Image Process.}, vol.~33, pp. 4543--4555, 2024.

\bibitem{zs2018}
Z.~Zou and Z.~Shi, ``Random access memories: {A} new paradigm for target detection in high resolution aerial remote sensing images,'' \emph{{IEEE} Trans. Image Process.}, vol.~27, no.~3, pp. 1100--1111, 2018.

\bibitem{ZSZ2022}
R.~Zhao, Z.~Shi, and Z.~Zou, ``High-resolution remote sensing image captioning based on structured attention,'' \emph{{IEEE} Trans. Geosci. Remote. Sens.}, vol.~60, pp. 1--14, 2022.

\bibitem{ZCZ2024}
H.~Zhang, H.~Chen, C.~Zhou, K.~Chen, C.~Liu, Z.~Zou, and Z.~Shi, ``Bifa: Remote sensing image change detection with bitemporal feature alignment,'' \emph{{IEEE} Trans. Geosci. Remote. Sens.}, vol.~62, pp. 1--17, 2024.

\bibitem{SLG2021}
Y.~Sun, L.~Lei, D.~Guan, and G.~Kuang, ``Iterative robust graph for unsupervised change detection of heterogeneous remote sensing images,'' \emph{{IEEE} Trans. Image Process.}, vol.~30, pp. 6277--6291, 2021.

\bibitem{TMD2020}
R.~Touati, M.~Mignotte, and M.~Dahmane, ``Multimodal change detection in remote sensing images using an unsupervised pixel pairwise-based markov random field model,'' \emph{{IEEE} Trans. Image Process.}, vol.~29, pp. 757--767, 2020.

\bibitem{LLM2018}
Z.~Liu, G.~Li, G.~Mercier, Y.~He, and Q.~Pan, ``Change detection in heterogenous remote sensing images via homogeneous pixel transformation,'' \emph{{IEEE} Trans. Image Process.}, vol.~27, no.~4, pp. 1822--1834, 2018.

\bibitem{ZGY2024}
Q.~Zhou, J.~Gao, Y.~Yuan, and Q.~Wang, ``Single-stream extractor network with contrastive pre-training for remote-sensing change captioning,'' \emph{{IEEE} Trans. Geosci. Remote. Sens.}, vol.~62, pp. 1--14, 2024.

\bibitem{CG2023}
S.~Chang and P.~Ghamisi, ``Changes to captions: An attentive network for remote sensing change captioning,'' \emph{{IEEE} Trans. Image Process.}, vol.~32, pp. 6047--6060, 2023.

\bibitem{LCC2024}
C.~Liu, K.~Chen, B.~Chen, H.~Zhang, Z.~Zou, and Z.~Shi, ``Rscama: Remote sensing image change captioning with state space model,'' \emph{{IEEE} Geosci. Remote. Sens. Lett.}, vol.~21, pp. 1--5, 2024.

\bibitem{SBC2024}
D.~Sun, Y.~Bao, and X.~Cao, ``A lightweight transformer for remote sensing image change captioning,'' \emph{arXiv}, vol. abs/2405.06598, 2024.

\bibitem{LZC2023}
C.~Liu, R.~Zhao, J.~Chen, Z.~Qi, Z.~Zou, and Z.~Shi, ``A decoupling paradigm with prompt learning for remote sensing image change captioning,'' \emph{{IEEE} Trans. Geosci. Remote. Sens.}, vol.~61, pp. 1--18, 2023.

\bibitem{ZWJ2024}
C.~Zhao, Y.~Wang, X.~Jiang, Y.~Shen, K.~Song, D.~Li, and D.~Miao, ``Learning domain invariant prompt for vision-language models,'' \emph{{IEEE} Trans. Image Process.}, vol.~33, pp. 1348--1360, 2024.

\bibitem{chatgpt}
\BIBentryALTinterwordspacing
OpenAI, ``Introducing chatgpt,'' Feb. 2023. [Online]. Available: \url{https://openai.com/blog/chatgpt}
\BIBentrySTDinterwordspacing

\bibitem{gpt4}
\BIBentryALTinterwordspacing
J.~Achiam, S.~Adler, S.~Agarwal, L.~Ahmad, I.~Akkaya, F.~L. Aleman, D.~Almeida, J.~Altenschmidt, S.~Altman, S.~Anadkat \emph{et~al.}, ``{GPT-4} technical report,'' \emph{arXiv}, vol. abs/2303.08774, 2023. [Online]. Available: \url{https://doi.org/10.48550/arXiv.2303.08774}
\BIBentrySTDinterwordspacing

\bibitem{GGZ2024}
P.~Gao, S.~Geng, R.~Zhang, T.~Ma, R.~Fang, Y.~Zhang, H.~Li, and Y.~Qiao, ``Clip-adapter: Better vision-language models with feature adapters,'' \emph{Int. J. Comput. Vis.}, vol. 132, no.~2, pp. 581--595, 2024.

\bibitem{CYC2024}
X.~Chen, J.~Yang, S.~Chen, L.~Wang, M.~Jiang, and Q.~Zhao, ``Every problem, every step, all in focus: Learning to solve vision-language problems with integrated attention,'' \emph{{IEEE} Trans. Pattern Anal. Mach. Intell.}, vol.~46, no.~7, pp. 4720--4735, 2024.

\bibitem{HWL2024}
C.~Han, C.~Wu, M.~Hu, J.~Li, and H.~Chen, ``C2f-semicd: {A} coarse-to-fine semi-supervised change detection method based on consistency regularization in high-resolution remote sensing images,'' \emph{{IEEE} Trans. Geosci. Remote. Sens.}, vol.~62, pp. 1--21, 2024.

\bibitem{LYQ2023}
C.~Liu, J.~Yang, Z.~Qi, Z.~Zou, and Z.~Shi, ``Progressive scale-aware network for remote sensing image change captioning,'' in \emph{Proc. {IEEE} Int. Geosci. Remote. Sens. Symp.}, 2023, pp. 6668--6671.

\bibitem{DSA2018}
R.~Caye~Daudt, B.~Le~Saux, and A.~Boulch, ``Fully convolutional siamese networks for change detection,'' in \emph{{IEEE} Int. Conf. Image Process.}, 2018, pp. 4063--4067.

\bibitem{HWG2023}
C.~Han, C.~Wu, H.~Guo, M.~Hu, J.~Li, and H.~Chen, ``Change guiding network: Incorporating change prior to guide change detection in remote sensing imagery,'' \emph{IEEE J. Sel. Top. Appl. Earth Obs. Remote Sens.}, vol.~16, pp. 8395--8407, 2023.

\bibitem{ZYT2020}
C.~Zhang, P.~Yue, D.~Tapete, L.~Jiang, B.~Shangguan, L.~Huang, and G.~Liu, ``A deeply supervised image fusion network for change detection in high resolution bi-temporal remote sensing images,'' \emph{ISPRS J. Photogramm. Remote Sens.}, vol. 166, pp. 183--200, 2020.

\bibitem{FLS2021}
S.~Fang, K.~Li, J.~Shao, and Z.~Li, ``Snunet-cd: A densely connected siamese network for change detection of vhr images,'' \emph{{IEEE} Geosci. Remote. Sens. Lett.}, vol.~19, pp. 1--5, 2022.

\bibitem{LYZ2023}
M.~Lin, G.~Yang, and H.~Zhang, ``Transition is a process: Pair-to-video change detection networks for very high resolution remote sensing images,'' \emph{{IEEE} Trans. Image Process.}, vol.~32, pp. 57--71, 2023.

\bibitem{JWW2023}
B.~Jiang, Z.~Wang, X.~Wang, Z.~Zhang, L.~Chen, X.~Wang, and B.~Luo, ``Vct: Visual change transformer for remote sensing image change detection,'' \emph{{IEEE} Trans. Geosci. Remote. Sens.}, vol.~61, pp. 1--14, 2023.

\bibitem{CQS2022}
H.~Chen, Z.~Qi, and Z.~Shi, ``Remote sensing image change detection with transformers,'' \emph{{IEEE} Trans. Geosci. Remote. Sens.}, vol.~60, pp. 1--14, 2022.

\bibitem{DZG2024}
L.~Ding, J.~Zhang, H.~Guo, K.~Zhang, B.~Liu, and L.~Bruzzone, ``Joint spatio-temporal modeling for semantic change detection in remote sensing images,'' \emph{{IEEE} Trans. Geosci. Remote. Sens.}, vol.~62, pp. 1--14, 2024.

\bibitem{LZD2022}
Q.~Li, R.~Zhong, X.~Du, and Y.~Du, ``Transunetcd: A hybrid transformer network for change detection in optical remote-sensing images,'' \emph{{IEEE} Trans. Geosci. Remote. Sens.}, vol.~60, pp. 1--19, 2022.

\bibitem{TZM2023}
X.~Tang, T.~Zhang, J.~Ma, X.~Zhang, F.~Liu, and L.~Jiao, ``Wnet: W-shaped hierarchical network for remote-sensing image change detection,'' \emph{{IEEE} Trans. Geosci. Remote. Sens.}, vol.~61, pp. 1--14, 2023.

\bibitem{HT2018}
H.~Jhamtani and T.~Berg{-}Kirkpatrick, ``Learning to describe differences between pairs of similar images,'' in \emph{Proc. Conf. Empir. Methods Nat. Lang. Process.}, 2018, pp. 4024--4034.

\bibitem{PDR2019}
D.~H. Park, T.~Darrell, and A.~Rohrbach, ``Robust change captioning,'' in \emph{Proc. {IEEE/CVF} Int. Conf. Comput. Vis.}, 2019, pp. 4623--4632.

\bibitem{SYG2020}
X.~Shi, X.~Yang, J.~Gu, S.~R. Joty, and J.~Cai, ``Finding it at another side: {A} viewpoint-adapted matching encoder for change captioning,'' in \emph{Proc. Eur. Conf. Comput. Vis.}, 2020, pp. 574--590.

\bibitem{HW2021}
M.~Hosseinzadeh and Y.~Wang, ``Image change captioning by learning from an auxiliary task,'' in \emph{Proc. {IEEE/CVF} Conf. Comput. Vis. Pattern Recognit.}, 2021, pp. 2725--2734.

\bibitem{ZMS2022}
D.~Zheng, F.~Meng, Q.~Si, H.~Fan, Z.~Xu, J.~Zhou, F.~Feng, and X.~Wang, ``Visual dialog for spotting the differences between pairs of similar images,'' in \emph{Proc. ACM Int. Conf. Multimedia}, 2022, pp. 5698--5709.

\bibitem{ZWW2024}
X.~Zhang, H.~Wen, J.~Wu, P.~Qin, L.~Nie \emph{et~al.}, ``Differential-perceptive and retrieval-augmented mllm for change captioning,'' in \emph{Proc. ACM Int. Conf. Multimedia}, 2024.

\bibitem{KKL2021}
H.~Kim, J.~Kim, H.~Lee, H.~Park, and G.~Kim, ``Viewpoint-agnostic change captioning with cycle consistency,'' in \emph{Proc. {IEEE/CVF} Int. Conf. Comput. Vis.}, 2021, pp. 2075--2084.

\bibitem{CHS2021}
S.~Chouaf, G.~Hoxha, Y.~Smara, and F.~Melgani, ``Captioning changes in bi-temporal remote sensing images,'' in \emph{Proc. {IEEE} Int. Geosci. Remote. Sens. Symp.}, 2021, pp. 2891--2894.

\bibitem{HCM2022}
G.~Hoxha, S.~Chouaf, F.~Melgani, and Y.~Smara, ``Change captioning: A new paradigm for multitemporal remote sensing image analysis,'' \emph{{IEEE} Trans. Geosci. Remote. Sens.}, vol.~60, pp. 1--14, 2022.

\bibitem{LZC2022}
C.~Liu, R.~Zhao, H.~Chen, Z.~Zou, and Z.~Shi, ``Remote sensing image change captioning with dual-branch transformers: A new method and a large scale dataset,'' \emph{{IEEE} Trans. Geosci. Remote. Sens.}, vol.~60, pp. 1--20, 2022.

\bibitem{PJM2024}
W.~Peng, P.~Jian, Z.~Mao, and Y.~Zhao, ``Change captioning for satellite images time series,'' \emph{{IEEE} Geosci. Remote. Sens. Lett.}, vol.~21, pp. 1--5, 2024.

\bibitem{LLW2023}
H.~Liu, C.~Li, Q.~Wu, and Y.~J. Lee, ``Visual instruction tuning,'' in \emph{Proc. Adv. Neural Inf. Process. Syst.}, 2023.

\bibitem{LLL2024}
H.~Liu, C.~Li, Y.~Li, and Y.~J. Lee, ``Improved baselines with visual instruction tuning,'' in \emph{Proc. {IEEE/CVF} Conf. Comput. Vis. Pattern Recognit.}, June 2024, pp. 26\,296--26\,306.

\bibitem{XCY2024}
Z.~Xiao, Y.~Chen, J.~Yao, L.~Zhang, Z.~Liu, Z.~Wu, X.~Yu, Y.~Pan, L.~Zhao, C.~Ma, X.~Liu, W.~Liu, X.~Li, Y.~Yuan, D.~Shen, D.~Zhu, D.~Yao, T.~Liu, and X.~Jiang, ``Instruction-vit: Multi-modal prompts for instruction learning in vision transformer,'' \emph{Inf. Fusion}, vol. 104, p. 102204, 2024.

\bibitem{LTC2024}
X.~Lai, Z.~Tian, Y.~Chen, Y.~Li, Y.~Yuan, S.~Liu, and J.~Jia, ``Lisa: Reasoning segmentation via large language model,'' in \emph{Proc. {IEEE/CVF} Conf. Comput. Vis. Pattern Recognit.}, June 2024, pp. 9579--9589.

\bibitem{WCC2023}
W.~Wang, Z.~Chen, X.~Chen, J.~Wu, X.~Zhu, G.~Zeng, P.~Luo, T.~Lu, J.~Zhou, Y.~Qiao, and J.~Dai, ``Visionllm: Large language model is also an open-ended decoder for vision-centric tasks,'' in \emph{Proc. Adv. Neural Inf. Process. Syst.}, 2023.

\bibitem{WSL2024}
W.~Wang, M.~Shi, Q.~Li, W.~Wang, Z.~Huang, L.~Xing, Z.~Chen, H.~Li, X.~Zhu, Z.~Cao, Y.~Chen, T.~Lu, J.~Dai, and Y.~Qiao, ``The all-seeing project: Towards panoptic visual recognition and understanding of the open world,'' in \emph{Proc. Int. Conf. Learn. Represent.}, 2024.

\bibitem{DFZ2024}
R.~Dang, J.~Feng, H.~Zhang, C.~Ge, L.~Song, L.~Gong, C.~Liu, Q.~Chen, F.~Zhu, R.~Zhao, and Y.~Song, ``Instructdet: Diversifying referring object detection with generalized instructions,'' in \emph{Proc. Int. Conf. Learn. Represent.}, 2024.

\bibitem{ZSC2023}
S.~Zhang, P.~Sun, S.~Chen, M.~Xiao, W.~Shao, W.~Zhang, K.~Chen, and P.~Luo, ``Gpt4roi: Instruction tuning large language model on region-of-interest,'' \emph{arXiv}, vol. abs/2307.03601, 2023.

\bibitem{CLD2024}
D.~Chen, J.~Liu, W.~Dai, and B.~Wang, ``Visual instruction tuning with polite flamingo,'' in \emph{Proc. {AAAI} Conf. Artif. Intell.}, 2024, pp. 17\,745--17\,753.

\bibitem{BBY2023}
J.~Bai, S.~Bai, S.~Yang, S.~Wang, S.~Tan, P.~Wang, J.~Lin, C.~Zhou, and J.~Zhou, ``Qwen-vl: {A} frontier large vision-language model with versatile abilities,'' \emph{arXiv}, vol. abs/2308.12966, 2023.

\bibitem{LCZ2024}
C.~Liu, K.~Chen, H.~Zhang, Z.~Qi, Z.~Zou, and Z.~Shi, ``Change-agent: Toward interactive comprehensive remote sensing change interpretation and analysis,'' \emph{{IEEE} Trans. Geosci. Remote. Sens.}, vol.~62, pp. 1--16, 2024.

\bibitem{CS2020}
H.~Chen and Z.~Shi, ``A spatial-temporal attention-based method and a new dataset for remote sensing image change detection,'' \emph{Remote Sens.}, vol.~12, no.~10, 2020.

\bibitem{LH2019}
I.~Loshchilov and F.~Hutter, ``Decoupled weight decay regularization,'' in \emph{Proc. Int. Conf. Learn. Represent.}, 2019.

\bibitem{ZZY2022}
Z.~Zhang, W.~Zhang, M.~Yan, X.~Gao, K.~Fu, and X.~Sun, ``Global visual feature and linguistic state guided attention for remote sensing image captioning,'' \emph{{IEEE} Trans. Geosci. Remote. Sens.}, vol.~60, pp. 1--16, 2022.

\bibitem{PRW2002}
K.~Papineni, S.~Roukos, T.~Ward, and W.~Zhu, ``Bleu: a method for automatic evaluation of machine translation,'' in \emph{Proc. Annu. Meet. Assoc. Comput. Linguist.}, 2002, pp. 311--318.

\bibitem{BL2005}
S.~Banerjee and A.~Lavie, ``{METEOR}: An automatic metric for {MT} evaluation with improved correlation with human judgments,'' in \emph{Proc. ACL Workshop Instrins. Extrins. Eval. Meas. Mach. Transl. Summ.}, 2005, pp. 65--72.

\bibitem{L2004}
C.-Y. Lin, ``Rouge: A package for automatic evaluation of summaries,'' in \emph{Text Summ. Branches Out}, 2004, pp. 74--81.

\bibitem{VZP2015}
R.~Vedantam, C.~L. Zitnick, and D.~Parikh, ``Cider: Consensus-based image description evaluation,'' in \emph{Proc. {IEEE/CVF} Conf. Comput. Vis. Pattern Recognit.}, 2015, pp. 4566--4575.

\bibitem{SEA2019}
S.~Liu, E.~Johns, and A.~J. Davison, ``End-to-end multi-task learning with attention,'' in \emph{Proc. {IEEE/CVF} Conf. Comput. Vis. Pattern Recognit.}\hskip 1em plus 0.5em minus 0.4em\relax Computer Vision Foundation / {IEEE}, 2019, pp. 1871--1880.

\bibitem{DFI2015}
A.~Dosovitskiy, P.~Fischer, E.~Ilg, P.~H{\"{a}}usser, C.~Hazirbas, V.~Golkov, P.~van~der Smagt, D.~Cremers, and T.~Brox, ``Flownet: Learning optical flow with convolutional networks,'' in \emph{Proc. {IEEE/CVF} Int. Conf. Comput. Vis.}, 2015, pp. 2758--2766.

\end{thebibliography}

\newpage

 




\vfill

\end{document}